\documentclass[letterpaper]{article} 
\usepackage{aaai25}
\usepackage{times}  
\usepackage{helvet}  
\usepackage{courier}  
\usepackage[hyphens]{url}  
\usepackage{graphicx} 
\usepackage{graphicx}
\usepackage{floatrow}
\usepackage{cite}
\usepackage{verbatim}
\usepackage{algorithmic}
\usepackage{algorithm}
\usepackage{array}
\usepackage{microtype}
\usepackage{multirow}
\usepackage{subcaption}
\usepackage{caption}
\usepackage{soul}
\usepackage{tikz}
\usepackage{bbding} 

\usepackage{microtype}
\usepackage{multirow}
\usepackage{graphicx}
\usepackage{subcaption}
\usepackage{svg}
\usepackage{color}
\usepackage{colortbl}
\usepackage{booktabs} 

\usepackage{amsmath}

\usepackage{amssymb}
\usepackage{mathtools}
\usepackage{amsthm}
\usepackage{subcaption}
\usepackage{color}
\usepackage{colortbl}
\usepackage{subcaption}
\usepackage{caption}

\usepackage{natbib}  
\usepackage{caption} 
\usepackage{algorithm}
\usepackage{algorithmic}

\usepackage{newfloat}
\usepackage{listings}
\DeclareCaptionStyle{ruled}{labelfont=normalfont,labelsep=colon,strut=off} 
\floatstyle{ruled}
\newfloat{listing}{tb}{lst}{}
\floatname{listing}{Listing}
\title{A Similarity Paradigm  Through Textual Regularization  Without Forgetting}
\author{
    Fangming Cui\textsuperscript{\rm 1},
    Jan Fong\textsuperscript{\rm 2},
    Rongfei Zeng\textsuperscript{\rm 3},
    Xinmei Tian\textsuperscript{\rm 4},
    Jun Yu\textsuperscript{\rm 5}\thanks{Corresponding author.}
}
\affiliations{
    \textsuperscript{\rm 1}Shanghai Jiao Tong University\\
    \textsuperscript{\rm 2}Hong Kong Baptist University\\
    \textsuperscript{\rm 3}Northeastern University\\
     \textsuperscript{\rm 4}University of Science and Technology of China\\
      \textsuperscript{\rm 5}Harbin Institute of Technology (Shenzhen)\\

    cuifangming@sjtu.edu.cn, zengrf@swc.neu.edu.cn, xinmei@ustc.edu.cn, yujun@hit.edu.cn\\
    
}

\usepackage{bibentry}

\begin{document}

\maketitle

\begin{abstract}
Prompt learning has emerged as a promising method for adapting pre-trained visual-language models (VLMs) to a range of downstream tasks. While optimizing the context can be effective for improving performance on specific tasks, it can often lead to poor generalization performance on unseen classes or datasets sampled from different distributions. It may be attributed to the fact that textual prompts tend to overfit downstream data distributions, leading to the forgetting of generalized knowledge derived from hand-crafted prompts.
In this paper, we propose a novel method called Similarity Paradigm with Textual Regularization  (SPTR) for prompt learning without forgetting. SPTR is a two-pronged design based on hand-crafted prompts that is an inseparable framework.  1) To avoid forgetting general textual knowledge, we introduce the optimal transport as a textual regularization to finely ensure approximation with hand-crafted features and tuning textual features. 2) In order to continuously unleash the general ability of multiple hand-crafted prompts, we propose a similarity paradigm for natural alignment score and adversarial alignment score to improve model robustness for generalization. Both modules share a common objective in addressing generalization issues, aiming to maximize the generalization capability derived from multiple hand-crafted prompts.
Four representative tasks
(i.e.,  non-generalization few-shot learning, base-to-novel generalization, cross-dataset generalization,  domain generalization) across  11 datasets demonstrate that SPTR outperforms existing prompt learning methods.
\end{abstract}

\section{Introduction}
Large vision-language models (VLMs) like pre-trained 
CLIP~\cite{radford2021learning} has demonstrated remarkable generalization capabilities across a wide range of downstream tasks~\cite{bangalath2022bridging,zhang2021tip,cao2023few}. 
By learning to associate textual descriptions with corresponding visual content, these models can perform tasks such as image classification, object detection, image generation, and more~\cite{cao2023break,xie2023videotrack,li2024sparseformer,zhang2022predict,cao2024neural,shi2024foodfusion,liu2024decoupling,liu2023fsi,liu2024motion}.
\begin{table}[t]
\small
 \centering
		\begin{tabular}{lcc} 
			\toprule 
			 Method&   4 tasks (Avg.) &    \\
			\midrule 
            
           CLIP (ICML2021)  & 67.69& $\Delta$\\
          
			CoOp (IJCV2022)& 68.31 & +0.62  \\  
   CoCoOp (CVPR2022) & 68.63 &  +0.94  \\ 
   KgCoOp (CVPR2023) & 70.71 &  +3.02 \\ 
   PLOT (ICLR2023) & 71.24  & +3.55\\
   MaPLe  (CVPR2023) &71.58 &  +3.89 \\ 
   PromptSRC (ICCV2023)& 72.09 & +4.40  \\
   TCP (CVPR2024)& 71.79 & +4.10 \\
   
  \midrule 
\rowcolor{gray!20}Ours (AAAI2025) &\textbf{72.89} & \textbf{\textcolor{black}{+5.20}}  \\              
			
            \bottomrule 
		\end{tabular}
\caption{We test our method on $4$ tasks: non-generalization few-shot learning, base-to-novel generalization, domain generalization, and cross-dataset evaluation.  Ours (AAAI2025) overall shows competitive results for average performance compared to the previous prompting methods. The symbol $\Delta$  represents the value that increases in comparison to CLIP.}	
\label{4task}
\end{table}

The generalization capabilities of these VLMs have been validated through their impressive performance on various benchmarks and tasks. They have shown strong performance in both seen and unseen classes, indicating their ability to generalize well to novel concepts and data distributions.
CLIP utilizes pre-defined text inputs or prompts during inference to generate classification weights. These prompts can include hand-engineered text, such as ``a photo of a [class]'', which guides the text encoder of CLIP.
As a zero-shot model, CLIP does not require task-specific fine-tuning on specific downstream tasks. This makes it highly versatile and capable of performing reasonably well on a wide range of tasks without any fine-tuning. It can leverage the alignment between two modalities learned during pre-training to make predictions and understand the semantic relationships between them.

Advanced techniques have been proposed to introduce learnable parameters for few-shot textual prompting to enhance the performance of pre-trained CLIP in downstream classification tasks and automate prompts. 
One significant method in this area is CoOp, introduced in the seminal work~\cite{zhou2022learning}. CoOp leverages the differentiable nature of neural networks to transform context words in a prompt into a set of learnable embeddings. By learning these textual embeddings of prompts, CoOp achieves substantial improvements over manually tuned prompts on various image recognition datasets, even with limited labeled images (few shots) available for training.

Although textual embedding tuning methods perform well on classification tasks for seen classes, they may not generalize effectively to unseen classes. As shown in Table~\ref{base2new} (left-top), the average performance of these textual embedding methods~\cite{zhou2022conditional,zhou2022learning, Yao_2023_CVPR,chen2022plot} on unseen classes is lower than that of hand-crafted CLIP ($74.22\%$). This suggests that textual embeddings may not adequately capture the semantic relationships and visual characteristics specific to the unseen classes. One possible reason for the degradation in performance for unseen classes is that textual prompts tend to overfit downstream data distributions~\cite{zhou2022conditional}. This overfitting can lead to the forgetting of hand-crafted general textual knowledge, which possesses strong generalization abilities.

To address the degradation~\cite{lu2022prompt, zhu2023prompt} in performance for unseen classes, we propose a 
novel method called Similarity Paradigm with Textual Regularization  \textbf{(SPTR)} for prompt learning. SPTR is a two-pronged design based on multiple hand-crafted prompts that is an inseparable prompting framework. 
We attempt to awaken and maintain pre-trained general knowledge, as general knowledge has excellent generalization ability. We use multiple hand-crafted prompts to create multiple pre-trained text features, fully leveraging the generalized capabilities of the CLIP. 
To bring the tuning features closer to the multiple pre-trained features, we apply the optimal transport as a textual regularization to finely ensure approximation with hand-crafted features of pre-trained CLIP and tuning textual features that avoid forgetting essential general textual knowledge.  
The optimal transport can finely calculate the distance between two distributions under the form of multiple sampling, which achieves fine-grained matching across two distributions. 
In order to continuously unleash the general ability of multiple hand-crafted prompts, we propose a similarity paradigm for natural alignment score and
adversarial alignment score to improve model robustness for generalization. 

As shown in Table~\ref{4task}, SPTR shows competitive results for average performance compared to the representative prompting methods, it demonstrates the robustness of SPTR in different types of classification tasks. 
Our main contributions can be summarized as follows:

\begin{itemize}
 \item  To avoid forgetting essential general textual knowledge, we propose to introduce the optimal transport plan as a textual regularization to finely ensure approximation with multiple hand-crafted features and tuning textual features.

\item Based on multiple hand-crafted prompts, we propose a novel similarity paradigm for natural alignment score and adversarial alignment score to improve generalization.

\item Extensive experiments show that SPTR performs favorably well on $4$ types of representative tasks across $11$ datasets compared to the existing prompt learning methods, achieving state-of-the-art performance.   
\end{itemize}

\section{Methodology}

\subsection{Preliminaries}
\noindent \textbf{Adversarial Training.}
Adversarial attacks~\cite{zhang2021causaladv,tang2022virtual,zhang2020principal} refer to the creation of maliciously crafted inputs, called adversarial examples, that are designed to deceive machine learning models~\cite{jia2022exploring, miyato2016adversarial,jia2020robust}. A common method~\cite{madry2017towards} with crafting adversarial examples for VLMs involves seeking a perturbation $\mathbf{r}$ for an input natural image $\mathbf{x}$ with label $y$ to maximize the dissimilarity, typically cosine dissimilarity. To optimize (minimize) the model after the maximum disturbance, the design of adversarial training for a tuning loss can be formalized as follows: 
\begin{equation}
\min _{\theta} \max _{\|\mathbf{r}\|_{p} \leq \epsilon} \mathcal{L}_{\text {EE}}(\mathbf{x}+ \mathbf{r}, y ;  \theta),
\end{equation}
where  $\|\cdot\|_p$ denotes the $\ell_p$-norm, $\theta$ denotes the tuning parameters, and $\epsilon$ controls the strength of adversarial perturbations. This design involves training models on adversarial examples generated to maximize prediction errors.

\begin{figure*}[ht]
\centerline{\includegraphics[scale=0.44]{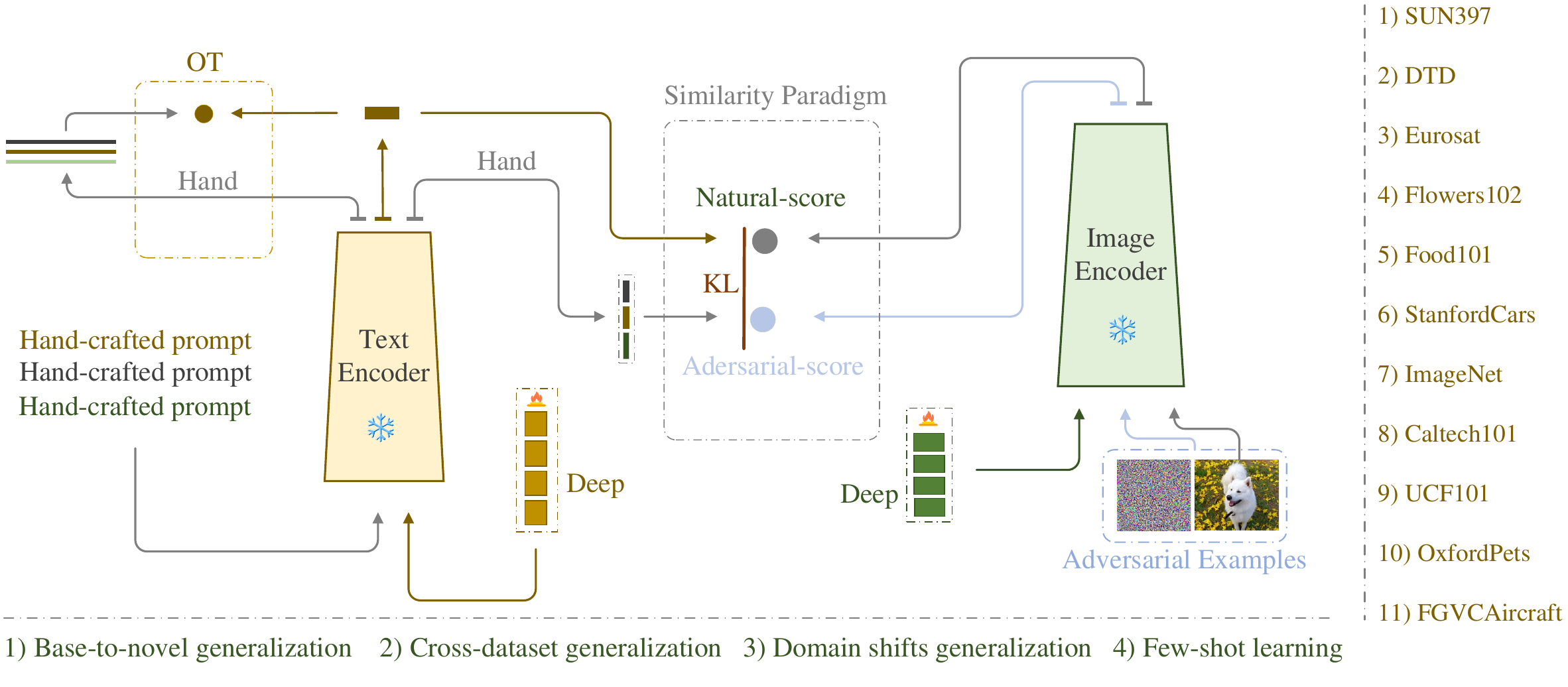}}
\caption{We propose a novel  similarity paradigm
with a textual regularization of the OT for prompting called SPTR. SPTR performs favorably well on $4$ types of representative tasks across $11$ datasets, achieving state-of-the-art performance. } 
\label{figframework}
\end{figure*}  

\begin{figure*}[ht]
\centerline{\includegraphics[scale=0.3]{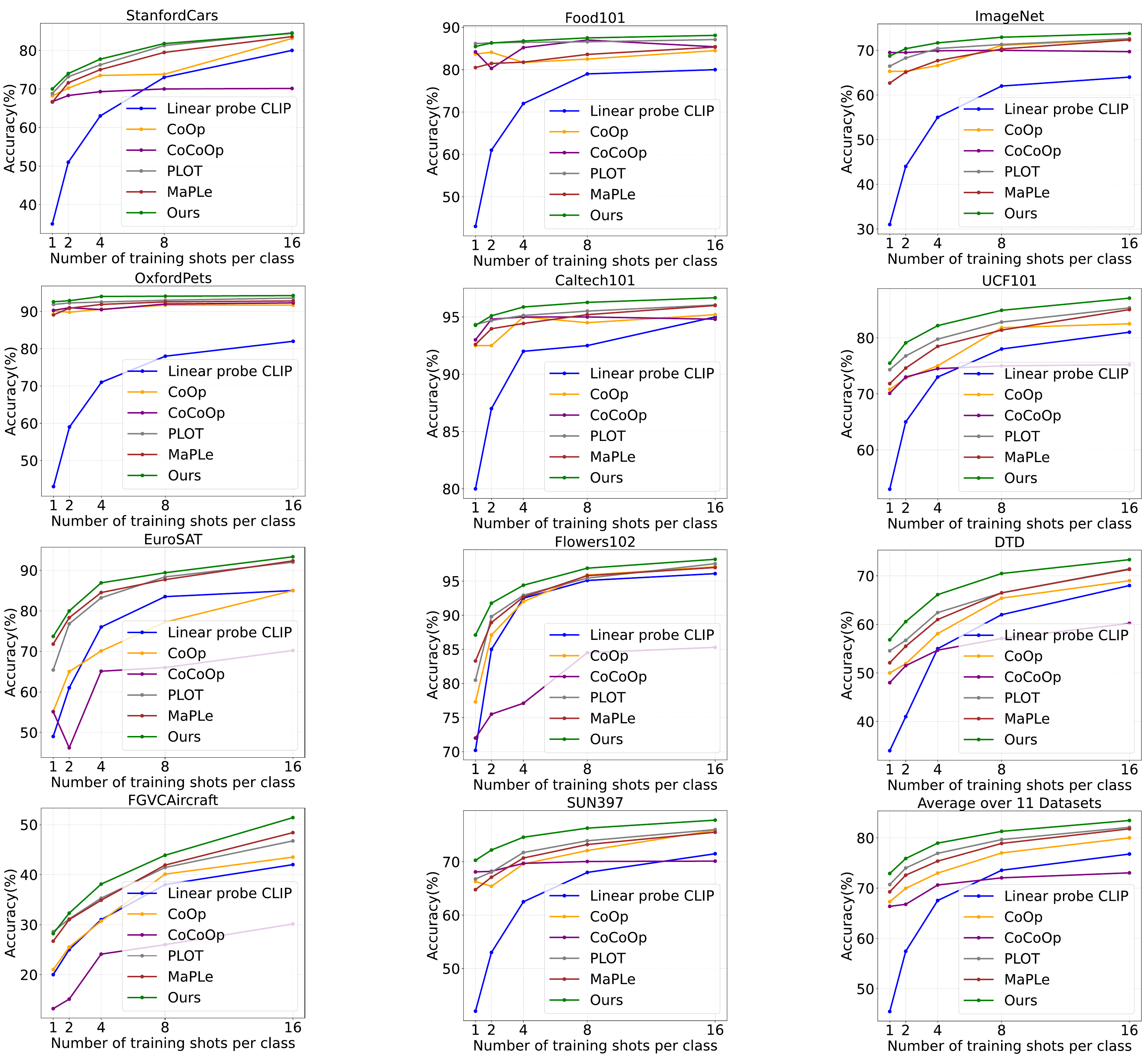}}
\caption{Non-generalization few-shot learning for all shots ($K = 1,2,4,8,16$).  
}
\label{few}
\end{figure*}  

\begin{table*}[ht]
\vskip 0.15in
\begin{center}
\begin{small}
\caption{Base-to-novel generalization experiments. `*’ denote
the performance obtained by our re-implementation.} 
\label{base2new}
\vskip -0.15in
\begin{floatrow}
\begin{subtable}{0.25\textwidth}

		\centering 
		\captionsetup{font={small}}
  \caption{Average over 11 datasets}
		
		\resizebox{1\textwidth}{!}{
  \renewcommand\arraystretch{1.0}
  
  \scalebox{0.5}{
  \setlength{\tabcolsep}{2pt}
			\begin{tabular}{lcc|c}
				\toprule
 & Base & Novel & HM \\
\midrule
CLIP  & 69.34& 74.22& 71.70 \\
CoOp    & 82.69& 63.22& 71.66 \\
Co & 80.47& 71.69& 75.83\\
Kg  & 80.73& 73.60& 77.00\\
PLOT   & 81.24   & 72.98  &  76.89 \\
MaPLe    & 82.28& 75.14& 78.55 \\
SRC*   & 84.12& 75.02& 79.31\\
TCP   & 84.13& 75.36& 79.51        \\
\midrule 
 \rowcolor{gray!20} SPTR  & \textbf{\textcolor{black}{84.85}}& \textbf{\textcolor{black}{76.76}}& \textbf{\textcolor{black}{80.61}}         \\
\bottomrule

			\end{tabular}
   }
		}
		
 \end{subtable}

 \hspace{-19pt}
	\hfill
   
 \begin{subtable}{0.25\textwidth}
		\centering
  \captionsetup{font={small}}
        \caption{ImageNet}

		\resizebox{1\textwidth}{!}{
  \renewcommand\arraystretch{1.0}
  \scalebox{0.5}{
  \setlength{\tabcolsep}{2pt}
			\begin{tabular}{lcc|c}
				\toprule
 & Base & Novel & HM \\
\midrule
CLIP  & 72.43& 68.14& 70.22 \\
CoOp    &  76.47&  67.88& 71.92 \\
Co & 75.98& 70.43& 73.10\\
Kg  & 75.83& 69.96& 72.78\\
PLOT   & 75.33   &  70.48 &  72.83 \\
MaPLe    &76.66&70.54&  73.47 \\
SRC*   &77.60& 70.73& 74.01        \\
TCP   &77.27& 69.87& 73.38        \\
\midrule 
\rowcolor{gray!20}SPTR  & \textbf{77.60}& \textbf{71.75}&   \textbf{74.56}      \\
\bottomrule
			\end{tabular}
   }
		}
		
\end{subtable}
\hspace{-19pt}
\hfill
\begin{subtable}{0.25\textwidth}
		\centering
  \captionsetup{font={small}}
        \caption{Caltech101}
		\resizebox{1\textwidth}{!}{
  \renewcommand\arraystretch{1.0}
 \scalebox{0.5}{
 \setlength{\tabcolsep}{2pt}
			\begin{tabular}{lcc|c}
				\toprule
 & Base & Novel & HM \\
\midrule
CLIP  & 96.84& 94.00& 95.40 \\
CoOp    & 98.00& 89.81&  93.73 \\
Co &  97.96&  93.81&  95.84\\
Kg  & 97.72& \textbf{94.39}& 96.03\\
PLOT   &  97.86  & 93.99  &  95.92 \\
MaPLe    & 97.74& 94.36& 96.02 \\
SRC*   &98.10& 93.90& 95.97        \\
TCP   &\textbf{98.23}& 95.33& \textbf{96.42}        \\
\midrule 
\rowcolor{gray!20}SPTR  & 98.01& 93.83&    95.87      \\
\bottomrule
			\end{tabular}
   }
		}
		
 \end{subtable}
 
\hfill
\hspace{-19pt}
\begin{subtable}{0.25\textwidth}


		\centering
  \captionsetup{font={small}}
        \caption{OxfordPets}

		\resizebox{1\textwidth}{!}{
  \renewcommand\arraystretch{1.0}
 \scalebox{0.5}{
 \setlength{\tabcolsep}{2pt}
			\begin{tabular}{lcc|c}
				\toprule
 & Base & Novel & HM \\
\midrule
CLIP  & 91.17& 97.26& 94.12 \\
CoOp    & 93.67& 95.29& 94.47 \\
Co & 95.20&  97.69&96.43\\
Kg  & 94.65& \textbf{98.10}& 96.38\\
PLOT   &  \textbf{95.7}  & 98.09  &  \textbf{96.89} \\
MaPLe    & 95.43& 97.76& 96.58 \\
SRC*   & 95.50&97.40&  96.44        \\
TCP   & 94.67&97.20&  95.92        \\
\midrule 
\rowcolor{gray!20}SPTR  & 94.85& 97.10&     95.97     \\
\bottomrule

			\end{tabular}
   }
		}
		
 \end{subtable} 
\end{floatrow}
 \begin{floatrow}

\begin{subtable}{0.25\textwidth}
		\centering
  \captionsetup{font={small}}
        \caption{EuroSAT}
		\resizebox{1\textwidth}{!}{
  \renewcommand\arraystretch{1.0}
  \scalebox{0.5}{
  \setlength{\tabcolsep}{2pt}
			\begin{tabular}{lcc|c}
				\toprule
 & Base & Novel & HM \\
\midrule
CLIP  & 56.48& 64.05& 60.03 \\
CoOp    & 92.19& 54.74&  68.69\\
Co &  87.49& 60.04& 71.21\\
Kg  & 85.64& 64.34& 73.48\\
PLOT   &  90.2  & 63.5  & 74.54  \\
MaPLe    & \textbf{94.07}&  73.23& \textbf{82.35} \\
SRC*   & 92.40& 68.43& 78.63        \\
TCP   & 91.63& \textbf{74.73}& 82.32        \\
\midrule 
\rowcolor{gray!20}SPTR  & 93.15& 72.80&      81.73    \\
\bottomrule
			\end{tabular}
   }
		}
		
\end{subtable}
\hspace{-19pt}
\hfill
\begin{subtable}{0.25\textwidth}
		\centering
  \captionsetup{font={small}}
        \caption{UCF101}

		\resizebox{1\textwidth}{!}{
  \renewcommand\arraystretch{1.0}
  \scalebox{0.5}{
  \setlength{\tabcolsep}{2pt}
			\begin{tabular}{lcc|c}
				\toprule
 & Base & Novel & HM \\
\midrule
CLIP  & 70.53&77.50& 73.85 \\
CoOp    & 84.69&  56.05& 67.46 \\
Co & 82.33&  73.45& 77.64\\
Kg  & 82.89& 76.67& 79.65\\
PLOT   &  82.56  &  75.56 &  78.92 \\
MaPLe    &  83.00& 78.66& 80.77 \\
SRC*   & 86.93&78.33& 82.41        \\
TCP   & 87.13&80.77& 83.83        \\
\midrule 
\rowcolor{gray!20}SPTR  & \textbf{88.17}& \textbf{81.00}&     \textbf{84.47}     \\
\bottomrule
			\end{tabular}
   }
		}
		
 \end{subtable}

 \hspace{-19pt}
	\hfill
\begin{subtable}{0.25\textwidth}
		\centering
  \captionsetup{font={small}}
        \caption{StanfordCars}

		\resizebox{1\textwidth}{!}{
  \renewcommand\arraystretch{1.0}
  \scalebox{0.5}{
  \setlength{\tabcolsep}{2pt}
			\begin{tabular}{lcc|c}
				\toprule
 & Base & Novel & HM \\
\midrule
CLIP  & 63.37& 74.89& 68.65 \\
CoOp    & 78.12& 60.40& 68.13 \\
Co &  70.49&  73.59&  72.01\\
Kg  & 71.76& 75.04& 73.36\\
PLOT   &  71.5  &  73.77 & 72.62  \\
MaPLe    & 72.94&  74.00& 73.47 \\
SRC*   &78.40& 74.73& 75.52        \\
TCP   &80.80& 74.13& 77.32        \\
\midrule 
\rowcolor{gray!20}SPTR  & \textbf{81.85}& \textbf{75.43}&     \textbf{78.53}     \\
\bottomrule
			\end{tabular}
   }
		}
		
\end{subtable}
\hfill
\hspace{-19pt}
\begin{subtable}{0.25\textwidth}
		\centering
  \captionsetup{font={small}}
        \caption{Flowers102}
		
		\resizebox{1\textwidth}{!}{
  \renewcommand\arraystretch{1.0}
  \scalebox{0.5}{
  \setlength{\tabcolsep}{2pt}
			\begin{tabular}{lcc|c}
				\toprule
 & Base & Novel & HM \\
\midrule
CLIP  & 72.08& 77.80& 74.83 \\
CoOp    &  97.60& 59.67& 74.06 \\
Co &  94.87& 71.75& 81.71\\
Kg  & 95.00& 74.73& 83.65\\
PLOT   &  95.10  & 72.20  & 82.10  \\
MaPLe    & 95.92& 72.46&  82.56 \\
SRC*   & 97.90&  76.77& 86.06       \\
TCP   & 97.73&  75.57& 85.23       \\
\midrule 
\rowcolor{gray!20}SPTR  & \textbf{98.56}& \textbf{77.59}&    \textbf{86.86}      \\
\bottomrule
			\end{tabular}
   }
		}
		
 \end{subtable}
\end{floatrow}
 \begin{floatrow}

\begin{subtable}{0.25\textwidth}


		\centering
  \captionsetup{font={small}}
        \caption{Food101}
		\resizebox{1\textwidth}{!}{
  \renewcommand\arraystretch{1.0}
  \scalebox{0.5}{
  \setlength{\tabcolsep}{2pt}
			\begin{tabular}{lcc|c}
				\toprule
 & Base & Novel & HM \\
\midrule
CLIP  & 90.10& 91.22& 90.66 \\
CoOp    &  88.33& 82.26&85.19 \\
Co & 90.70& 91.29& 90.99\\
Kg  & 90.5& 91.7& 91.09\\
PLOT   &  90.98  & 91.54  & 91.28  \\
MaPLe    & 90.71& 92.05& 91.38 \\
SRC*   &90.63& 91.50& 91.06        \\
TCP   &90.57& 91.37& 90.97        \\
\midrule 
\rowcolor{gray!20}SPTR  & \textbf{91.11}& \textbf{92.74}&    \textbf{91.94}      \\
\bottomrule

			\end{tabular}
   }
		}
		
 \end{subtable}

 \hspace{-19pt}
	\hfill
\begin{subtable}{0.25\textwidth}
		\centering
  \captionsetup{font={small}}
        \caption{FGVCAircraft}
		\resizebox{1\textwidth}{!}{
  \renewcommand\arraystretch{1.0}
  \scalebox{0.5}{
  \setlength{\tabcolsep}{2pt}
			\begin{tabular}{lcc|c}
				\toprule
 & Base & Novel & HM \\
\midrule
CLIP  & 27.19&36.29& 31.09 \\
CoOp    & 40.44& 22.30& 28.75 \\
Co &  33.41& 23.71& 27.74\\
Kg  & 36.21& 33.55& 34.83\\
PLOT   &  35.6  &  28.5 &  31.66 \\
MaPLe    & 37.44& 35.61& 36.50 \\
SRC*   & 42.30& 36.97& 39.46        \\
TCP   & 41.97& 34.43& 37.83        \\
\midrule 
\rowcolor{gray!20}SPTR  & \textbf{44.26}& \textbf{40.18}&     \textbf{42.09}     \\
\bottomrule
			\end{tabular}
   }
		}
		
\end{subtable}
\hspace{-19pt}
\hfill
\begin{subtable}{0.25\textwidth}
		\centering
  \captionsetup{font={small}}
        \caption{SUN397}		
		\resizebox{1\textwidth}{!}{
  \renewcommand\arraystretch{1.0}
  \scalebox{0.5}{
  \setlength{\tabcolsep}{2pt}
			\begin{tabular}{lcc|c}
				\toprule
 & Base & Novel & HM \\
\midrule
CLIP  & 69.36& 75.35 & 72.23 \\
CoOp    &  80.60& 65.89& 72.51 \\
Co & 79.74& 76.86&  78.27\\
Kg  & 80.29& 76.53& 78.36\\
PLOT   & 79.96   &  77.33 &  78.64 \\
MaPLe    & 80.82& 78.70& 79.75 \\
SRC*   & 82.83& 79.00& 80.87        \\
TCP   & 82.63& 78.20& 80.35        \\
\midrule 
\rowcolor{gray!20}SPTR  & \textbf{82.56}& \textbf{79.26}&    \textbf{80.90}      \\
\bottomrule
			\end{tabular}
   }
		}
		
 \end{subtable}
\hfill
\hspace{-19pt}
\begin{subtable}{0.25\textwidth}


		\centering
  \captionsetup{font={small}}
        \caption{DTD}		
		\resizebox{1\textwidth}{!}{
  \renewcommand\arraystretch{1.0}
  \scalebox{0.5}{
  \setlength{\tabcolsep}{2pt}
			\begin{tabular}{lcc|c}
				\toprule
 & Base & Novel & HM \\
\midrule
CLIP  & 53.24& 59.90& 56.37 \\
CoOp    &  79.44&  41.18& 54.24 \\
Co & 77.01&  56.00&  64.85\\
Kg  & 77.55& 54.99& 64.35\\
PLOT   &  78.9  & 57.9  & 66.8  \\
MaPLe    & 80.36&  59.18& 68.16 \\
SRC*   & 82.60& 57.50& 67.80        \\
TCP   & 82.77& 58.07& 68.25        \\
\midrule 
\rowcolor{gray!20}SPTR  & \textbf{83.35}& \textbf{62.58}&    \textbf{71.50}      \\
\bottomrule

			\end{tabular}
   }
		}
		
 \end{subtable}
\end{floatrow}

\end{small}
\end{center}

\end{table*}

\begin{table*}[h]
\small
\caption{ Cross-dataset benchmark evaluation. The symbol $\Delta$  represents the value that increases in comparison to CLIP.}

\centering
\setlength{\tabcolsep}{3pt}
\begin{tabular}{lccccccccccccc}
\toprule 
          & \textbf{Source} & \multicolumn{9}{c}{\textbf{Target}}          
          \\ \cmidrule(r){2-2} \cmidrule(r){3-12}
          & \multicolumn{1}{c}{\raisebox{-2.5ex}{\rotatebox[origin=c]{90}{\centering ImageNet} }}     
          & \multicolumn{1}{c}{\raisebox{-2.5ex}{\rotatebox[origin=c]{90}{Caltech101}}}    
          &\multicolumn{1}{c}{\raisebox{-2.5ex}{\rotatebox[origin=c]{90}{OxfordPets}}}     
          & \multicolumn{1}{c}{\raisebox{-2.5ex}{\rotatebox[origin=c]{90}{StanfordCars}}}  
          &\multicolumn{1}{c}{\raisebox{-2.5ex}{\rotatebox[origin=c]{90}{Flowers102}}}     
          & \multicolumn{1}{c}{\raisebox{-2.5ex}{\rotatebox[origin=c]{90}{Food101}}}       
          & \multicolumn{1}{c}{\raisebox{-2.5ex}{\rotatebox[origin=c]{90}{Aircraft}}}       
          &\multicolumn{1}{c}{\raisebox{-2.5ex}{\rotatebox[origin=c]{90}{SUN397} }}        
          & \multicolumn{1}{c}{\raisebox{-2.5ex}{\rotatebox[origin=c]{90}{DTD} }}          
          & \multicolumn{1}{c}{\raisebox{-2.5ex}{\rotatebox[origin=c]{90}{EuroSAT}}}       
          & \multicolumn{1}{c}{\raisebox{-2.5ex}{\rotatebox[origin=c]{90}{UCF101} }}       
          & \multicolumn{1}{c}{\raisebox{-2.5ex}{\rotatebox[origin=c]{90}{{Average}} }} 
          & \multicolumn{1}{c}{\raisebox{-2.5ex}{\rotatebox[origin=c]{90}{\textit{ }} }}
          \\ 
          
          \midrule 

CLIP      & 66.72  & 92.94          & 89.07          & 65.29          & 71.30          & 86.11          & 24.87          & 62.62          & 44.56          & 47.69          & 66.77          & 65.12        &   $\Delta$    \\

CoOp      & 71.51  & 93.70          & 89.14          & 64.51          & 68.71          & 85.30          & 18.47          & 64.15          & 41.92          & 46.39          & 66.55          & 63.88        &   \textcolor{black}{-2.3}    \\

Co   & 71.02           & 94.43          & 90.14          & 65.32 & 71.88          & 86.06          & 22.94          & \textbf{67.36}          & 45.73& 45.37          & 68.21          & 65.74       &  +0.6     \\ 

Kg   & 70.66           & 93.92        & 89.83          & 65.41 & 70.01          & \textbf{86.36}          & 22.51          & 66.16          & 46.35& 46.04          & 68.50          & 65.51        &  +0.4  \\
PLOT   & 70.15           & \textbf{94.60}        & 90.23          & 65.41 & 71.97          & 86.32          & 22.87          & 67.22          & 44.99& 46.57          & 68.32          & 65.85  & +0.7         \\

MaPLe   & 70.72           & 93.53         & 90.49          & 65.57 & \textbf{72.23}          & 86.20          & \textbf{24.74}          & 67.01          & 46.49& 48.06          & 68.69          & 66.30        &   +1.2    \\

SRC   & 71.27           & 93.60         & 90.25          & 65.70 & 70.25          & 86.15          & 23.90          & 67.10          & 46.87 & 45.50          & 68.75          & 65.81    & +0.7     \\

TCP   & 71.40           & 93.97        & 91.25          & 64.69 & 71.21          & 86.69          & 23.45          & 67.15          & 44.35& 51.45          & 68.73          & 66.26  &+1.2          \\

 \midrule
 
 \rowcolor{gray!20}Ours & 70.05           & 94.45 & \textbf{90.70} & \textbf{65.83}          &72.22 & 86.13 & 23.60 &67.00 & \textbf{49.13}          & \textbf{51.33} & \textbf{69.10} & \textbf{66.95}  &\textbf{\textcolor{black}{+1.8}}\\ 
\bottomrule
\end{tabular}

\label{cross}
\end{table*} 

\noindent \textbf{Pre-Trained CLIP.}
Our method is based on a pre-trained vision-language model, hand-crafted CLIP, which is a zero-shot learning method~\cite{wortsman2022robust, Li_2024_CVPR}.  The pre-trained CLIP has two encoders: 
a frozen text encoder and a frozen image encoder, which separately map a hand-crafted textual input ${p}$ and training images $\mathbf{x}$ into a shared feature space through transformer blocks. 
In the pre-training stage of CLIP, text encoder $\mathcal{F}_{t}(\cdot)$ and image encoder $\mathcal{F}_{v}(\cdot)$ undergo simultaneous training on extensive datasets containing text-image pairs. To achieve this, a contrastive loss function  $\mathcal{L}_ {\text{CE}}$ is utilized to maximize the cosine similarity between text-image pairs while minimizing the cosine similarity between unmatched pairs within the feature space:
\begin{align}
\mathcal{L}_ {\text{CE}}  = -\log \frac{\exp \left\{ \operatorname{sim} \left(\mathcal{F}_{t}\left({p}_{y}\right), \mathcal{F}_{v}(\mathbf{x})\right) / \tau\right\}}{\sum_{{p}_{i} \in \mathcal{P}} \exp \left\{\operatorname{sim} \left(\mathcal{F}_{t}\left({p}_{i}\right), \mathcal{F}_{v}(\mathbf{x})\right) / \tau\right\}},
\end{align}
where  $y \in \mathcal{Y}$  is the label of $ \mathbf{x} \in \mathcal{X}$, $\mathcal{P}=\left\{{p}_{i}\right\}_{i=1}^{K} $  presents the set of  $K$  hand-crafted prompts,  $\operatorname{sim} (\cdot, \cdot)$  stands for cosine similarity between textual features and visual features, and  $\tau$  denotes a temperature parameter. To this end, the classifier consists of  $K$  textual features derived from hand-crafted prompts  $\mathcal{P}=\left\{{p}_{i}\right\}_{i=1}^{K}$, where the prompt  ${p}_{i}$  for the  $i$-th class may have the form of ``a picture of a'', ``a bad photo of a'', ``a photo of many'', etc~\cite{radford2021learning}.

\noindent \textbf{Textual Prompting.}
Recent studies suggest that fine-tuning prompts for specific images may outperform the use of manually crafted prompts. More precisely, the class name is preserved as prior knowledge to guarantee that the acquired prompts can construct a classifier, whereas the word (also known as context) embeddings of prompts are treated as adjustable parameters. The learnable words in the above prompt are initialized using  ``a photo of a [classname]".  In the inference phase, the prompts equipped with the acquired context can generate textual features for classification purposes.

\noindent \textbf{Textual Regularization.}
The textual regularization techniques~\cite{Yao_2023_CVPR,khattak2023self,yao2023tcp} refer to the regularization of pre-trained and fine-tuned features limited to the text branch through metric function to prevent forgetting essential general knowledge. The key concept for this component is to awaken and sustain pre-trained general knowledge~\cite{2023understanding}, given its exceptional generalization capabilities, which address the degradation in performance for unseen classes.  

For the first time, we employ optimal transport (OT)~\cite{villani2009optimal} as a textual regularization, thereby preventing the loss of essential general textual knowledge. The optimal transport facilitates the computation of distances between two distributions (multiple pre-trained textual features and single-tuning textual features).
Unlike traditional distance metrics~\cite{yao2024tcp, Yao_2023_CVPR}, such as L1 and MSE, the optimal transport adopts a detailed matching strategy with multiple hand-crafted prompts to measure distances between distributions, which is different from single hand-crafted prompts~\cite{yao2024tcp, Yao_2023_CVPR}. 
Specifically, let $\boldsymbol{T}^{sets}_{tun}$ denote the tuning textual feature sets, and let $\boldsymbol{T}^{sets}$ denote the pre-trained hand-crafted textual feature sets.  
Further, we can obtain $M$ tuning local textual feature belongs to $\boldsymbol{T}^{sets}_{tun}$, and $N$ pre-trained local textual features with $N$ hand-crafted prompts belong to $\boldsymbol{T}^{sets}$. Note that $M=1$ because the learnable textual prompt is a single prompt, which is different from PLOT with multiple learnable prompts.  Following the OT algorithm, our method learns the plan by minimizing the following distance $Dis$ to push $\boldsymbol{T}^{sets}$ to $\boldsymbol{T}^{sets}_{tun}$ for fine-grained alignment.
To this end, we can obtain the textual regularization distance $Dis$.

\subsection{Proposed Similarity  Paradigm}
In order to continuously unleash the general ability of multiple hand-crafted prompts, we propose a similarity paradigm for natural alignment score and adversarial alignment score to improve model robustness for generalization. When training models for vision and language tasks, it is crucial to consider both natural and adversarial scenarios. By narrowing the distribution gap between natural text-image similarity and adversarial text-image similarity, the model is encouraged to make more consistent and robust predictions in the presence of adversarial inputs. 

Inspired by distributional robust optimization, we propose to perform distributional exploration for the image distribution, which would identify the worst-case distribution for tuning prompts. We optimize the learnable visual prompt  $\mathcal{P}_{v}$   and learnable textual prompt $\mathcal{P}_{t}$ using the adversarial images as the identified worst-case distribution for robustness improvement. The adversarial examples are obtained by adding adversarial perturbations $\mathbf{r}$ to natural images $\mathbf{x}$.  Following previous adversarial attack methods, we apply the projected gradient descent (PGD) algorithm~\cite{madry2017towards} to search for adversarial perturbation $\mathbf{r}$ iteratively:
\begin{equation}
\mathbf{r}^{(t+1)}= {\prod}_{\mathcal{C}(\mathbf{r}, \epsilon, p)}
(\mathbf{r}^t + \mu \delta^t_\mathbf{r}),  \label{adv_example}
\end{equation}
where ${\prod}$ designates a projection function, $\mathcal{C}(\mathbf{r}, \epsilon, p)$ stands for the $\ell_p$-norm ball centered at the embedding $\mathbf{r}$ through a radius $\epsilon$, $\mu$ is the step size, $\delta^t$ represents the update direction at the $t$-th iteration. To solve the constrained optimization problem, we can derive the update direction as follows:
\begin{equation}
    \delta^t = 
    \epsilon
    \textit{sign}(\nabla_\mathbf{x}\mathcal{L}_{\mathrm{EE}}(\mathbf{x}, y, \mathbf{r}))
    \frac{\left| \nabla_\mathbf{x}\mathcal{L}_{\mathrm{EE}}(\mathbf{x}, y, \mathbf{r})\right|^{q-1}}
    {(\Vert
    \nabla_\mathbf{x}\mathcal{L}_{\mathrm{EE}}(\mathbf{x}, y, \mathbf{r})
    \Vert
    ^q_q)^{1/p}},
\end{equation}
where $\textit{sign}(\cdot)$ designates the sign function, $\mathcal{L}_{\mathrm{EE}}(\cdot)$ designates a loss, $q$ is the dual of $p$.

To this end, we propose a similarity paradigm for minimizing the Kullback-Leibler (KL) divergence between natural Vision-Language (V-L) alignment score (cosine similarity) and adversarial V-L alignment score, as shown in  Figure~\ref{figframework}.
The similarity paradigm aims to mitigate the potential vulnerability to robustness issues stemming from disparities in generalizing natural samples.
Specifically, let $\boldsymbol{v}_{n}$ denote the output visual features of natural images, $\boldsymbol{v}_{adv} $ denote the output visual features of adversarial examples,   $\boldsymbol{t}_{tun}$  denote tuning textual feature. Further, we let $\boldsymbol{t}_{hand}$ denote the compressed hand-crafted feature, which is the average of $N$ templates hand-crafted features~\cite{radford2021learning}, leading to learning more diverse generalized knowledge from the frozen CLIP.  The $\operatorname{sim} (\cdot, \cdot)$  stands for cosine score (logit) between visual features and textual features. The objective of the Similarity Paradigm (SP)  can be stated as:
\begin{equation}
    \mathcal{L}_{\text {SP }}=\mathcal{KL}\big(\operatorname{sim}\left(\boldsymbol{t}_{tun}, \boldsymbol{v}_{n}\right), \operatorname{sim}(\boldsymbol{t}_{hand}, \boldsymbol{v}_{adv})\big).
\end{equation}
Accordingly, the final objective  $\mathcal{L}_{\text {total }}$    of SPTR with a  textual hyper-parameter $\alpha$ can be formulated as follows:
\begin{equation}
\mathcal{L}_{\text {total}}= \mathcal{L}_{\text {CE}}+  \mathcal{L}_{\text {SP}} + \alpha Dis,
\end{equation}
where  $\mathcal{L}_{\text {CE}}$  represents nature alignment of cross-entropy loss.
The $Dis$ avoids forgetting essential general textual knowledge of hand-crafted CLIP  having a strong generalization ability. Based on multiple hand-crafted prompts, $\mathcal{L}_{\text {SP }}$ as a KL constraint for natural alignment score and adversarial  alignment score that mitigates  the potential failure in robustness
caused by discrepancies in natural generalization.

\section{Experiments}
\subsection{Experimental Settings}

\noindent \textbf{Baselines.}
We compare the experimental performance based on  ViT-B/16 CLIP with these methods: CLIP~\cite{radford2021learning}, CoOp~\cite{zhou2022learning}, CoCoOp (Co)~\cite{zhou2022conditional}, KgCoOp (Kg)~\cite{Yao_2023_CVPR}, PLOT~\cite{chen2022plot},
MaPLe~\cite{khattak2023maple}, PromptSRC (SRC)~\cite{khattak2023self} and TCP~\cite{yao2024tcp}. Differing from the textual single-modal regularization of our method, PLOT employs the optimal transport plan in the multi-modal regularization for alignment with visual and textual prompting branches.

\noindent \textbf{Datasets.}
The datasets cover multiple scenes
including ImageNet~\cite{deng2009imagenet} and Caltech101~\cite{fei2004learning}, OxfordPets~\cite{parkhi2012cats}, StanfordCars~\cite{krause20133d},
Flowers102~\cite{nilsback2008automated}, Food101~\cite{bossard2014food}, FGVCAircraft ~\cite{maji2013fine}, SUN397~\cite{xiao2010sun} UCF101~\cite{soomro2012ucf101}, DTD~\cite{cimpoi2014describing} and EuroSAT ~\cite{helber2019eurosat}. We
use ImageNetA~\cite{hendrycks2021natural}, ImageNet-R~\cite{hendrycks2021many}, ImageNet-Sketch~\cite{wang2019learning} and ImageNetV2~\cite{recht2019imagenet} for domain generalization.

\noindent \textbf{Implementation Details.}
We set the embeddings length to $4$. We train  $50$ epochs for few-shot learning tasks and $20$ epochs for other tasks. The learning rate is 0.0025 via SGD optimizer on a single GPU.  We use the ViT-B/16 model-based CLIP and set $\alpha$ to $0.3$. For domain generalization and cross-dataset evaluation, we train the ImageNet source model on all classes in the first $3$ layers of encoders. 
For the base-to-novel settings and few-shot learning, we set learning depth to $9$. We set $N$ to $60$, which is consistent with pre-trained CLIP. Attacks are generated with a perturbation boundary $\epsilon$ = $1/255$.

\begin{table}[h]
\small
\caption{Domain generalization experiments. The symbol $\Delta$  represents the value that increases in comparison to CLIP. `*' denotes the performance obtained by our re-implementation.}

\centering
\setlength{\tabcolsep}{3pt}
\begin{tabular}{lccccccc}
\toprule 
          & \textbf{Source} & \multicolumn{4}{c}{\textbf{Target}}          
          \\ \cmidrule(r){2-2} \cmidrule(r){3-6}
          
          & \multicolumn{1}{c}{\centering ImageNet}      
          & \multicolumn{1}{c}{\centering -V2}   
          &\multicolumn{1}{c}{\centering -S}   
          & \multicolumn{1}{c}{\centering -A}   
          &\multicolumn{1}{c}{\centering -R}            
          & \multicolumn{1}{c}{\centering Avg.} 
          & \multicolumn{1}{c}{\centering }
          \\ 
          \midrule 
CLIP      & 66.73  & 60.83          & 46.15
          & 47.77
          & 73.96
          & 57.18
          & $\Delta$
           \\
CoOp      & 71.51  & 64.2
          & 47.99
          & 49.71
          & 75.21
          & 59.28
          &+2.1
           \\
Co   & 71.02           & 64.07
          & 48.75
          & 50.63 & 76.18
          & 59.91
          & +2.8
          \\ 

Kg    & 70.66           & 64.1
          & 48.97
          & 50.69 & 76.7
          & 60.11
          & +2.9
          \\           
PLOT    & 70.15           & 64.17
          & 49.15
          & 50.83 & 76.5
          & 60.16
          & +2.9
          \\
MaPLe  & 70.72           & 64.07
          & 49.15
          & 50.9 &76.98
          & 60.27
          & +3.1
         \\

SRC*  & 71.27           & 64.35
          & \textbf{49.55}
          & 50.90 &76.80
          & 60.40
          & +3.2
         \\ 
         
TCP    &  71.40           & 64.05
          & 49.10
          & 50.55 &  77.8
          & 60.37
          & +3.2
          \\

 \midrule
 
 \rowcolor{gray!20}Ours &   70.05         & \textbf{64.40} & 48.78 & \textbf{51.30}
          &\textbf{77.9} & \textbf{60.59} &\textbf{\textcolor{black}{+3.4}}\\ 
\bottomrule
\end{tabular}

\label{domain}
\end{table} 

\subsection{Experimental Results}  

\noindent \textbf{Few-Shot Learning.}
The evaluation of the model is performed at different $K$-shot levels per class, encompassing values of 1, 2, 4, 8, and 16. In Figure~\ref{few}, our method consistently delivers enhancements across all few-shot settings when compared to existing methods. 
SPTR achieves improved performance in non-generalization classification, even when training resources are limited ($K = 2$). 

\noindent \textbf{Base-to-Novel Generalization.}
The datasets are split into base and novel classes.
The model is trained only on the base classes in the $16$ shots setting and evaluated on base and novel classes. In Table~\ref{base2new}, SPTR provides the best-averaged results of $84.85\%$, $76.76\%$, and $80.61\%$ on the base classes, novel classes, and harmonic mean, respectively.

\noindent \textbf{Domain Generalization.}
We train our model using the ImageNet in $16$ shots and evaluate its performance directly on $4$ different variants of the ImageNet. In Table~\ref{domain}, 
SPTR outperforms all existing methods on target datasets,
with an overall highest average accuracy of $60.59\%$. Compared with TCP (CVPR2024), SPTR shows improved performance in ImageNet-A with a large margin. 

\noindent \textbf{Cross-Dataset Generalization.}
We evaluate our method on $10$ unseen datasets.  In Table~\ref{cross}, in comparison with TCP (CVPR2024), SPTR  achieves better generalization on average for $10$ datasets. This
indicates that SPTR favors better generalization for a wide range of unseen datasets.

\section{Further Analysis}

\subsection{Training and Inference Compute Cost Analysis}
In Table~\ref{cost}, we show the computed cost analysis ~\cite{wang2021spgnet,yang2022diversity} of SPTR
and compare it with prompt learning methods. Compared to MaPLe, SPTR has fewer parameters. CoCoOp is significantly slower, due to the single output design of the image encoder. CoCoOp transfers the output features of the image encoder to the text embeddings, which is more inference time consumption.   Due to the presence of adversarial designs in SPTR, the inference speed is relatively slow~\cite{madry2017towards}.

\begin{table}[ht]
\small
		\caption{The computational cost comparison using SUN397
dataset. Training time for all methods is calculated for 10 epochs
on a single GPU.}		
		\centering
   
		\begin{tabular}{lccc} 
			\toprule 
			 Method& Params   & Train time (min) & HM     \\
			\midrule 
            CoOp& 2048   & 10.88&71.65     \\             
			Co&  35360  & 39.53&75.83    \\
            Kg & 2048& 11.25 & 77.00\\
            PLOT & 8192& 10.85& 76.48\\
            MaPLe& 3.55 M  & 10.58& 79.68   \\
            SRC & 5120 &  13.13 & 79.97 \\
            TCP & 2048& 11.85 & 80.11\\
            \midrule 
            \rowcolor{gray!20}SPTR (Ours)& 5120  &  14.50& \textbf{80.31}    \\                
			\bottomrule 
		\end{tabular}
   
  \label{cost}
	\end{table}

\subsection{Embeddings Length and Learning Depth}
In Table~\ref{length}, our findings indicate that the performance reaches its peak when the length of embeddings is set to $4$ on base-to-novel generalization for HM.   In Table~\ref{depth}, we note that increasing the learning depth generally increases the performance. 
As the number of layers increases to $11$, the HM value decreases. It indicates excessive fine-tuning of the model, causing it to lose CLIP  generalization and the ability of V-L alignment~\cite{zhou2022conditional}. 

\begin{table}[ht]
\small
\centering
\caption{Analysis of embeddings length (HM).}

\setlength{\tabcolsep}{4pt}
\begin{tabular}{lccccc}
\toprule
          {Embeddings Length}          
          & {{\centering 1} }     
          & {{2}}    
          &\textbf{4}
          & {{6}}  
          &{{8}}

          \\ 
          
          \midrule 

Caltech101   & 95.13           & 95.31         & \textbf{95.87}          & 93.10 & 95.33                     \\

Food101 & 89.11           & 91.09         & \textbf{91.94}          & 88.19 & 90.92                      \\
\bottomrule 
\end{tabular}
\label{length}
\end{table} 

\begin{table}[ht]
\small
\centering
\caption{Analysis of learning depth (HM).}

\setlength{\tabcolsep}{4pt}
\begin{tabular}{lcccccc}
\toprule
          {{Learning Depth}}           
          & {{\centering 1} }     
          & {{3}}    
          &{5}     
          & {{7}}  
          &{\textbf{9}}    
          & {{11}}      
           
          \\ 
          
          \midrule 

Caltech101  & 91.83          & 94.12          & 95.11& 95.19          & \textbf{95.87}          & 95.33         \\

Food101  & 86.53          & 88.00          & 89.71& 91.01          & \textbf{91.94}          & 91.10              \\

\bottomrule 

\end{tabular}
\label{depth}
\end{table} 

\begin{table}[h]
\small
\centering
\caption{Applying to ViT instances for Avg. (11 datasets).}
\centering
\setlength{\tabcolsep}{2pt}
\begin{tabular}{lccc}
\toprule
          
          & {{\centering ViT-B/32} }  
          & {ViT-B/16} 
          & {{\textbf{EVA-CLIP}}}
          
          \\ 
          \midrule 

HM   & 78.55          & 80.61        & \textbf{83.12}        \\

\bottomrule 
\end{tabular}
\label{ViT}
\end{table} 

\subsection{Distance Measurement for Textual Branch}
The analysis in Table~\ref{textual} suggests that, among the various textual regularization techniques explored, the OT (ours) leads to the best performance on base-to-novel generalization for 11 datasets. Cosine similarity primarily focuses on the angular similarity between vectors, MSE and L1 loss are designed to measure the distance between individual data points, rather than the overall distribution of the data. The optimal transport can calculate the distance between two distributions with a fine-grained matching to align features.

\begin{table}[!]
\small
		\caption{Distance measurement for Avg. (11 datasets).}	
  \centering
 
 \setlength{\tabcolsep}{2pt}
 
		\begin{tabular}{lccc} 
			\toprule 
			 Design& Base Acc. & Novel Acc. &  HM     \\
            \midrule 
            MSE  & 84.81 & 74.19 &   79.17 \\
            L1 & 84.60 & 76.00 & 80.09 \\
            Cosine & 84.79 & 75.51 & 79.90   \\
            \midrule 
            \rowcolor{gray!20}OT (Ours) & \textbf{84.85} &\textbf{76.76}  &\textbf{80.61}    \\   
            \bottomrule 
		\end{tabular}
   
  \label{textual}
	\end{table}

\begin{table}[!]
\small
\centering
\caption{Ablation of perturbation boundary $\epsilon$.}
\centering
\setlength{\tabcolsep}{5pt}
\begin{tabular}{ccccccc}
\toprule
          & \multicolumn{3}{c}{Caltech101} & \multicolumn{3}{c}{Food101}          
          \\ \cmidrule(r){2-4} \cmidrule(r){5-7}
         $\epsilon$ & \multicolumn{1}{c}{{\centering \textbf{1/255}} }  
          & \multicolumn{1}{c}{{4/255 }}  
          & \multicolumn{1}{c}{{8/255 }}
          &\multicolumn{1}{c}{\textbf{1/255} }                 
          & \multicolumn{1}{c}{{4/255 }}      
           & \multicolumn{1}{c}{{8/255}}
          \\ 
          \midrule 

HM   & \textbf{95.87}           & 94.00         & 93.58 &\textbf{91.94}          & 90.88     &  89.45       \\
\bottomrule 

\end{tabular}
\label{epsilon}
\end{table}

\begin{table}[!]
\small
\centering
\caption{Ablation of textual regularization $\alpha$.}
\centering
\setlength{\tabcolsep}{5pt}
\begin{tabular}{ccccccc}
\toprule
& \multicolumn{3}{c}{Caltech101} &
\multicolumn{3}{c}{Food101}          
          \\ \cmidrule(r){2-4} \cmidrule(r){5-7}
          $\alpha$& \multicolumn{1}{c}{{\centering 0.1} }  
          & \multicolumn{1}{c}{\textbf{ 0.3}}  
          & \multicolumn{1}{c}{{0.5}}
          &\multicolumn{1}{c}{0.1}                 
          & \multicolumn{1}{c}{\textbf{0.3}}      
           & \multicolumn{1}{c}{{0.5}}
          \\ 
          \midrule 

HM  & 93.00           & \textbf{95.87}         & 95.18 & 88.22          & \textbf{91.94}     &  90.32       \\

\bottomrule 

\end{tabular}
\label{alpha}
\end{table}

\begin{table}[ht]
\small
		\caption{Analysis of our components for Avg. (11 datasets).}	
  \centering
 \setlength{\tabcolsep}{2pt}
		\begin{tabular}{lccc} 
			\toprule 
			 Component& Base Acc. & Novel Acc. &  HM     \\
            \midrule 
            1: Baseline model  & 84.32 & 71.38 &  77.34 \\
            2: + OT  & 84.39 & 74.90 & 79.36  \\
            \midrule 
            \rowcolor{gray!20}3: + OT + Similarity paradigm & \textbf{84.85} & \textbf{76.76} & \textbf{80.61}\\
            \bottomrule 
		\end{tabular} 
  \label{components}
	\end{table}

\subsection{Analysis of Perturbation Boundary and Textual $\alpha$}
In Table~\ref{epsilon} and Table~\ref{alpha}, our findings indicate that the performance reaches its peak when the perturbation boundary $\epsilon$ is set to $1/255$ and $\alpha$ to 0.3 on base-to-novel generalization.

\subsection{Applying to Other ViT Instances}
Our framework was discovered to be adaptable to various types of Vision Transformer (ViT) instances, such as EVA-CLIP~\cite{Fang_2023_CVPR} and ViT-B/32 in Table~\ref{ViT}. In the future, we will validate our approach on more ViT architectures.

\subsection{Analysis of Integrated Contribution}
We employ an independent V-L prompting framework as our baseline model (row-1). By enforcing the optimal transport (OT) for textual regularization based on row-1, row-2 is higher than row-1 on novel performance in Table~\ref{components}. This is because OT plays a crucial role in mitigating the forgetting of general textual knowledge with a fine-grained perspective. By integrating a similarity paradigm based on row-2, we observed that row-3 outperforms row-2 on novel classes while maintaining base class gains. It suggests that the similarity paradigm enhances the ability of generalized robustness.

\section{Related Work} 
\subsection{Prompt Learning for Vision-Language Models}
Recently, the strong generalization capability of CLIP has made it a foundation for many methods in adapting pre-trained Vision-Language models (VLMs) for downstream tasks. 
Prompt tuning is a widely used technique in VLMs for learning downstream tasks. 
The use of text prompts, which are instructions provided to the language branch of VLMs, is a common practice to enhance task understanding. Full fine-tuning~\cite{cao2024controlled} and linear probe are two common methods used to adapt VLMs to downstream tasks. 
Some methods~\cite{zhou2022learning,zhou2022conditional,yao2023tcp} fine-tune the CLIP model specifically for few-shot image recognition by optimizing a continuous set of token embeddings in the language branch. 
Moreover, some methods
constrain the prompts to contain the essential general knowledge and prior distribution learning~\cite{lu2022prompt}. 
In addition to multi-modal prompt tuning~\cite{lee2023multimodal,khattak2023maple,khattak2023self}, some methods~\cite{zhou2024few,li2024one} apply adversarial training perspective for prompt tuning to effectively align its V-L representations for pre-trained CLIP.

\section{Conclusion}
Prompt learning involves optimizing the embeddings for adaptation to downstream tasks. Advanced
prompt learning methods usually initialize and optimize the context, e.g., ``a photo of a []'', for downstream task adaptation. 
While optimizing the embeddings can enhance performance on specific tasks, it may result in subpar generalization performance on unseen classes or datasets drawn from diverse distributions. 
To mitigate this issue, we propose SPTR, which is a two-pronged design based on hand-crafted prompts. Four representative tasks
(i.e.,  non-generalization few-shot learning, base-to-novel generalization, cross-dataset generalization,  domain generalization) across  11 datasets demonstrate that SPTR outperforms existing prompt learning methods.

\section{Acknowledgments}
This work was supported by the  National Natural Science Foundation of China (62125201).

\bibliography{arxiv}

\appendix

\end{document}